\documentclass{article}

\PassOptionsToPackage{numbers, compress}{natbib}
    \usepackage[preprint]{neurips_2023}

\usepackage[utf8]{inputenc} % allow utf-8 input
\usepackage[T1]{fontenc}    % use 8-bit T1 fonts
\usepackage[pagebackref,breaklinks,colorlinks]{hyperref}
\usepackage{url}            % simple URL typesetting
\usepackage{booktabs}       % professional-quality tables
\usepackage{amsfonts}       % blackboard math symbols
\usepackage{nicefrac}       % compact symbols for 1/2, etc.
\usepackage{microtype}      % microtypography
\usepackage{xcolor}         % colors
\usepackage{CJKutf8}

\usepackage[nolist]{acronym}
\makeatletter
\newcommand*{\org@overidelabel}{}
\let\org@overridelabel\@verridelabel
\@ifpackagelater{acronym}{2015/03/21}{% v1.41
  \renewcommand*{\@verridelabel}[1]{%
    \@bsphack
    \protected@write\@auxout{}{\string\AC@undonewlabel{#1@cref}}%
    \org@overridelabel{#1}%
    \@esphack
  }%
}{% older versions
  \renewcommand*{\@verridelabel}[1]{%
    \@bsphack
    \protected@write\@auxout{}{\string\undonewlabel{#1@cref}}%
    \org@overridelabel{#1}%
    \@esphack
  }%
}
\makeatother

\usepackage{amsmath}
\usepackage{amsfonts}
\usepackage{amssymb}
\usepackage{wrapfig}
\usepackage{subcaption}
\usepackage{multirow}
 \usepackage{mathtools} 

\usepackage{verbatim}

\usepackage{anyfontsize}

\usepackage{microtype}
\usepackage{graphicx}
\usepackage{booktabs} % for professional tables
\usepackage{multirow}
\usepackage{amsmath,amssymb}
\usepackage{booktabs}
\usepackage{caption,subcaption}

\usepackage{xcolor}
\definecolor{mygreen}{HTML}{3cb44b}
\definecolor{skyblue}{HTML}{beffff}
\definecolor{lightgreen}{HTML}{90ee90}

\usepackage{color, colortbl}

\definecolor{emerald}{rgb}{0.31, 0.78, 0.37}

\usepackage{tcolorbox}
\usepackage{enumitem}
\setitemize{itemsep=10pt,topsep=0pt,parsep=0pt,partopsep=0pt}
\usepackage{colortbl}

\usepackage{xcolor}
\definecolor{mygreen}{HTML}{3cb44b}
\colorlet{myyellow}{green!10!orange!90!}
\makeatletter

\usepackage{tikz}
\usetikzlibrary{arrows,shapes,automata,backgrounds,fit,petri}
\usepackage{adjustbox}

\newcommand{\RN}[1]{%
	\textup{\lowercase\expandafter{\it \romannumeral#1}}%
}
\usepackage{tabu}
\newcommand{\ie}[0]{\emph{i.e., }}

\newcommand{\eg}[0]{\emph{e.g., }}

\newcommand{\etc}[0]{\emph{etc.}}

\newcommand{\beq}{\vspace{0mm}\begin{equation}}
\newcommand{\eeq}{\vspace{0mm}\end{equation}}
\newcommand{\beqs}{\vspace{0mm}\begin{eqnarray}}
\newcommand{\eeqs}{\vspace{0mm}\end{eqnarray}}
\newcommand{\barr}{\begin{array}}
\newcommand{\earr}{\end{array}}

\newcommand{\Xmat}[0]{{{\bf X}}}

\usepackage{color, colortbl}
\definecolor{Gray}{gray}{0.93}

\usepackage{lipsum}

\newcommand\blfootnote[1]{%
  \begingroup
  \renewcommand\thefootnote{}\footnote{#1}%
  \addtocounter{footnote}{-1}%
  \endgroup
}

\usepackage{pifont}% http://ctan.org/pkg/pifont

\usepackage{makecell}

\usepackage{xcolor,amsmath}
\usepackage[linesnumbered,ruled,vlined]{algorithm2e}
\DontPrintSemicolon

\usepackage{xcolor}
\definecolor{mygreen}{HTML}{3cb44b}

\SetKwComment{Comment}{\color{green!50!black}\# }{}

\newcommand{\var}{\texttt}

\SetKwProg{Function}{def}{:}{}

\SetKwProg{For}{for}{:}{}
\SetKwProg{If}{if}{:}{}
\newcommand{\VarSty}[1]{\textnormal{\ttfamily\color{blue!90!black}#1}\unskip}

\usepackage{cleveref}
\usepackage{alltt}
\tcbuselibrary{most}
\tcbset{
  aibox/.style={
    width=\textwidth,
    top=10pt,
    colback=white,
    colframe=black,
    colbacktitle=black,
    enhanced,
    center,
    attach boxed title to top left={yshift=-0.1in,xshift=0.15in},
    boxed title style={boxrule=0pt,colframe=white,},
  }
}
\newtcolorbox{AIbox}[2][]{aibox,title=#2,#1}
\newlength\savewidth\newcommand\shline{\noalign{\global\savewidth\arrayrulewidth
  \global\arrayrulewidth 1pt}\hline\noalign{\global\arrayrulewidth\savewidth}}
\newcommand{\tablestyle}[2]{\setlength{\tabcolsep}{#1}\renewcommand{\arraystretch}{#2}\centering\footnotesize}

\newcommand{\shortname}{LLaVA-Med}
\newcommand{\longname}{{\bf L}arge {\bf L}anguage {\bf a}nd {\bf V}ision {\bf A}ssistant for Bio{\bf Med}icine}

\newcommand{\eat}[1]{\ignorespaces}

\title{\shortname{}: Training a Large Language-and-Vision Assistant for Biomedicine in One Day}

\author{
\textbf{\normalsize{  Chunyuan Li$^{*}$, Cliff Wong$^{*}$, Sheng Zhang$^{*}$, Naoto Usuyama, Haotian Liu, Jianwei Yang }}\\
\textbf{\normalsize{ Tristan Naumann, Hoifung Poon, Jianfeng Gao }}  \vspace{1mm}\\
Microsoft\\
\url{https://aka.ms/llava-med}
}

\begin{document}

\maketitle

\begin{acronym}
    \acro{LM}{language model}
    \acro{LLM}{large language model}
    \acro{LMM}{large multimodal model}
    \acro{NLP}{natural language processing}
    \acro{VL}{vision-language}
    \acro{VQA}{visual question answering}
\end{acronym}

\begin{CJK*}{UTF8}{gbsn}

\begin{abstract}
Conversational generative AI has demonstrated remarkable promise for empowering biomedical practitioners, but current investigations focus on unimodal text.
Multimodal conversational AI has seen rapid progress by leveraging billions of image-text pairs from the public web, but such general-domain vision-language models still lack sophistication in understanding and conversing about biomedical images.
In this paper, we propose a cost-efficient approach for training a vision-language conversational assistant that can answer open-ended research questions of biomedical images.
The key idea is to leverage a large-scale, broad-coverage biomedical figure-caption dataset extracted from PubMed Central, use GPT-4 to self-instruct open-ended instruction-following data from the captions, and then fine-tune a large general-domain vision-language model using a novel curriculum learning method. Specifically, the model first learns to align biomedical vocabulary using the figure-caption pairs as is, then learns to master open-ended conversational semantics using GPT-4 generated instruction-following data, broadly mimicking how a layperson gradually acquires biomedical knowledge.
This enables us to train a \longname{} (\shortname{})\blfootnote{$^*$Equal Contribution} in less than 15 hours (with eight A100s). \shortname{} exhibits excellent multimodal conversational capability and can follow open-ended instruction to assist with inquiries about a biomedical image. 
On three standard biomedical visual question answering datasets, fine-tuning \shortname{} outperforms previous supervised state-of-the-art on certain metrics.
To facilitate biomedical multimodal research, we will release our instruction-following data and the \shortname{} model.
\end{abstract}

\section{Introduction}
\label{sec:introduction}

Parallel image-text data is abundantly available in the general domain, such as web images and their associated captions. Generative pretraining has proven effective to leverage this parallel data for self-supervised vision-language modeling, as demonstrated by multimodal GPT-4~\citep{gpt4} and open-sourced efforts such as LLaVA~\cite{liu2023visual}.
By instruction-tuning models to align with human intents based on multimodal inputs, the resulting \acp{LMM} exhibit strong zero-shot task completion performance on a variety of user-oriented vision-language tasks such as image understanding and reasoning, paving the way to develop general-purpose multimodal conversational assistants~\cite{askell2021general,li2022elevater,gan2022vision}.

While successful in the general domains, such \acp{LMM} are less effective for biomedical scenarios because biomedical image-text pairs are drastically different from general web content. As a result, general-domain visual assistants may behave like a layperson, who would refrain from answering biomedical questions, or worse, produce incorrect responses or complete hallucinations.
Much progress has been made in biomedical \ac{VQA}, but prior methods typically formulate the problem as classification (\eg among distinct answers observed in the training set) and are not well equipped for open-ended instruction-following. 
Consequently, although conversational generative AI has demonstrated great potential for biomedical applications~\cite{peterbook2023,nori2023,lee2023benefits}, current investigations are often limited to unimodal text. 

% {\bf TODO add ref nori2023: https://arxiv.org/abs/2303.13375 and peter's book}

In this paper, we present \longname{} (\shortname{}), a first attempt to extend multimodal instruction-tuning to the biomedical domain for end-to-end training of a biomedical multimodal conversational assistant. 
Domain-specific pretraining has been shown to be effective for biomedical \ac{NLP} applications~\cite{lee2020biobert,huang2019clinicalbert,gu2021domain,luo2022biogpt} and biomedical \ac{VL} tasks~\cite{johnson2019mimic,boecking2022making,shih2019augmenting,zhang2023large,eslami2023pubmedclip}.
Most recently, large-scale biomedical \ac{VL} learning has been made possible by the creation of PMC-15M~\cite{zhang2023large}, a broad-coverage dataset with 15 million biomedical image-text pairs extracted from PubMed Central\footnote{\url{https://www.ncbi.nlm.nih.gov/pmc/}}. This dataset is two orders of magnitude larger than the next largest public dataset, MIMIC-CXR~\cite{johnson2019mimic}, and covers a diverse image types.
Inspired by recent work in instruction-tuning~\cite{peng2023instruction,liu2023visual}, \shortname{} uses GPT-4 to generate diverse biomedical multimodal instruction-following data using image-text pairs from PMC-15M, and fine-tune a large biomedical-domain \ac{VL} model~\cite{liu2023visual} using a novel curriculum learning method. 

%Specifically, the model first learns to align biomedical vocabulary using the figure-caption pairs as is, then learns to master open-ended conversational semantics using GPT-4 generated instruction-following data, broadly mimicking how a layerperson gradually acquires biomedical knowledge. This enables us to train a \longname{} (\shortname{}) in less than 15 hours (with eight A100s). \shortname{} exhibits excellent multimodal conversational capability and can follow open-ended instruction to assist with inquiries about a biomedical image. On two standard biomedical visual question answering datasets, zero-shot \shortname{} outperforms previous supervised state of the art by a large margin. To facilitate biomedical multimodal research, we will release our instruction-following data and the \shortname{} model.

%Specifically, large-scale biomedical vision-language pre-training %becomes feasible due to the recent release of has been explored on PMC-15M~\cite{zhang2023large}, a dataset with 15 million biomedical image-text pairs, which is two orders of magnitude larger than the previous largest public dataset MIMIC-CXR~\cite{johnson2019mimic} and covers a diverse range of image types.

Specifically, our paper makes the following contributions:

\begin{itemize}[leftmargin=7.5mm]
\setlength{\itemsep}{2pt}
\item 
{\it Biomedical multimodal instruction-following data}. We present a novel data generation pipeline to create diverse (image, instruction, output) instances, by sampling biomedical image-text pairs from PMC-15M and using GPT-4 to create instructions from the text alone (which becomes the intended output). This requires zero manual annotations and creates an extremely diverse visual instruction-following dataset by piggybacking on PMC-15 that covers the full spectrum of research findings over biomedical images.
%We present a data curation pipeline to convert image-text pairs into instruction-following data using language-only GPT-4.
%This leads to the first dataset for instruction-tuning for LMMs as in previous released biomedical vision-language datasets, such as PMC-15M, texts only describe the content of their paired images. 
%It raises the challenge that  

\item
{\it \shortname{}}. We propose a novel curriculum learning method for adapting LLaVA~\cite{liu2023visual} to the biomedical domain using our self-generated biomedical multi-modal instruction-following dataset. Specifically, we first fine-tune LLaVA to align biomedical vocabulary using the image-text pairs as is (with the generic instruction that simply asks for a  description of the image). We then continue training the model using our self-generated instruction-following data to learn open-ended conversational semantics. 
In this way, we were able to train \shortname{} in less than 15 hours with eight A100s. Our empirical study validates the effectiveness of domain-specific instruction-tuning, and reveals best practice and interesting findings for adapting multimodal conversational assistant to high-value verticals. On well-established biomedical \ac{VQA} datasets,  fine-tuning \shortname{} often outperforms supervised state-of-the-art (SoTA).

%We have developed a domain-specific \ac{LMM}, \shortname{}, by adapting the general-domain \ac{LMM} LLaVA~\cite{liu2023visual} using the curated medical instruction-following dataset. Our empirical study validates the effectiveness of domain-specific instruction-tuning, and reveals practical tips and interesting findings when building visual assistant via model adaptation from general to specific domains. We also study \shortname{} on three well-established medical \ac{VQA} datasets, and show new state-of-the-art results on the close-set problems.

\item
{\it Open-source}. To facilitate research in biomedical multimodal learning, we will release the following assets to the public: the biomedical multimodal instruction-following dataset and the codebase for data generation and model training.
% , and a demo of our biomedical multimodal conversational assistant. 

\end{itemize}

\vspace{-3mm}
\section{Related Work}
\label{sec:related_work}
\vspace{-3mm}

% \paragraph{Biomedical multimodal learning} Discuss RoentGen (Text-to-Image Generation) and contrastive learning (BioViL ...). SEM / pix2pix?

\paragraph{Biomedical Chatbots.} Inspired by ChatGPT~\cite{chatgpt}/GPT-4~\cite{gpt4} and the success of open-sourced instruction-tuned \acp{LLM} in the general domain, several biomedical \ac{LLM} chatbots have been developed, including ChatDoctor~\cite{yunxiang2023chatdoctor}, Med-Alpaca~\cite{han2023medalpaca}, PMC-LLaMA~\cite{wu2023pmc}, Clinical Camel~\cite{clinical_camel}, DoctorGLM~\cite{xiong2023doctorglm}, and Huatuo~\cite{wang2023huatuo}. They are initialized with open-sourced \ac{LLM} and fine-tuned on  customized sets of biomedical instruction-following data. The resulting \acp{LLM} emerge with great potential to offer  assistance in a variety of biomedical-related fields/settings, such as understanding patients' needs and providing informed advice. 

% ChatDoctor~\cite{yunxiang2023chatdoctor} collected 700+ diseases and their corresponding symptoms, required medical tests, and recommended medications, from which  5K  doctor-patient conversations are generated. By fine-tuning LLMs using these tailored doctor-patient conversations, the resulting models emerge with great potential to understand patients' needs, provide informed advice, and offer valuable assistance in a variety of medical-related fields.

% MedAlpaca~\cite{han2023medalpaca} presents a dataset with over 160K entries, specifically crafted to fine-tune LLMs for effective medical applications.

% PMC-LLaMA~\cite{wu2023pmc} is fine-tuned on 4.8 million biomedical academic papers to gain medical knowledge, enhancing its capability in medical domain.

% Clinical Camel~\cite{clinical_camel} builds on the performance seen by fine-tuning LLaMa with a mixture of user-shared conversations and synthetic conversations designed to encode high-quality clinical data from curated clinical articles.

% DoctorGLM~\cite{xiong2023doctorglm}

% Huatuo~\cite{wang2023huatuo} is fine-tuned with Chinese medical instructions

To our knowledge, Visual Med-Alpaca~\cite{wu2023_visual-med_alpaca} is the only existing multimodal biomedical chatbot that accepts image inputs. Though Visual Med-Alpaca and the proposed \shortname{} share a similar input-output data format, they differ in key aspects:
$(i)$ {\it Model architectures.} \shortname{} is an end-to-end neural model and Visual Med-Alpaca is a system that connect multiple image captioning models with a \ac{LLM}, using a classifier to determine if or which biomedical captioning model is responsible for the image. The text prompt subsequently merges the converted visual information with the textual query, enabling Med-Alpaca to generate an appropriate response.
$(ii)$ {\it Biomedical instruction-following data.} While Visual Med-Alpaca is trained on 54K samples from limited biomedical subject domains, \shortname{} is trained a more diverse set.

% Multiple rounds of human filtering and editing are performed to refine the question-answer pairs, resulting in a high-quality instruction set comprising 54k data points. Next, we expand Med-Alpaca into Visual Med-Alpaca by connecting the textual model with "visual medical experts," which are specialized medical computer vision models. For instance, in radiology-domain applications, we train an in-house radiology image captioning model called Med-GIT (see later for details). When given an input image, a classifier determines if or which medical visual expert is responsible for the image. The designated medical expert then converts the image into a text prompt. The prompt manager subsequently merges the converted visual information with the textual query, enabling Med-Alpaca to generate an appropriate response.

\paragraph{Biomedical Visual Question Answering.}
An automated approach to building models that can answer questions based on biomedical images stands to support clinicians and patients.
% It is a valuable asset to have an automated approach/system that is able to answer questions based on biomedical images, giving insight to clinicians and patients. 
To describe existing biomedical \ac{VQA} methods, we make a distinction between discriminative and generative methods. 
For discriminative methods, \ac{VQA} is treated a classification problem: models make predictions from a predefined set of answers. While discriminative methods yield good performance, they deal with closed-set predictions~\cite{he2020pathvqa}, and require mitigation when a customized answer set is provided in at inference~\cite{li2022self,zhang2023large,eslami2023pubmedclip}. The discriminative formulation is suboptimal towards the goal of developing a general-purpose biomedical assistant that can answer open questions in the wild. To this end, generative methods have been developed to predict answers as a free-form text sequence~\cite{bazi2023vision,liu2023q2atransformer,van2023open}. Generative methods are more versatile because they naturally cast the close-set questions as as special case where candidate answers are in language instructions.

\paragraph{Model Architecture.}
\shortname{} is similar to prefix tuning of \acp{LM} in~\cite{van2023open} in that a new trainable module connects frozen image encoder and causal \ac{LM}. In~\cite{van2023open}, a three-layer MLP network is used to map the visual features into a visual prefix, and the pre-trained LM are GPT2-XL~\cite{radford2019language}, BioMedLM~\cite{venigalla2022biomedlm} and BioGPT~\cite{luo2022biogpt}, with size varying from 1.5B to 2.7B. By contrast, \shortname{} uses a linear projection and a 7B \ac{LM}~\cite{vicuna,touvron2023llama}. 
Most importantly, \cite{van2023open} only considers standard supervised fine-tuning and focuses efforts on exploring various modeling choices. Our main contributions instead comprise proposing a novel data generation method that uses GPT-4 to self-instruct biomedical multimodal instruction-following data using freely-available broad-coverage biomedical image-text pairs extracted from PubMed Central~\cite{zhang2023large}.
% {\bf DOUBLE CHECK - }

%More importantly, we emphasize that we undertake and advocate {\it data-centric} paradigm, while all existing methods are {\it model-centric}. Specifically, we simplify model architecture and training, and focus on iteratively improving the quality of instruct data, which represents first dataset of its kind in healthcare domains.

%{\bf TODO: key = zero-shot vs dataset-specific fine-tuning}
\begin{figure}[!ht]
\begin{AIbox}{Biomedical Visual Instruction-Following Example}

\begin{wrapfigure}{r}{0.3\textwidth}\vspace{-3pt}
\centering    \includegraphics[width=0.3\textwidth]{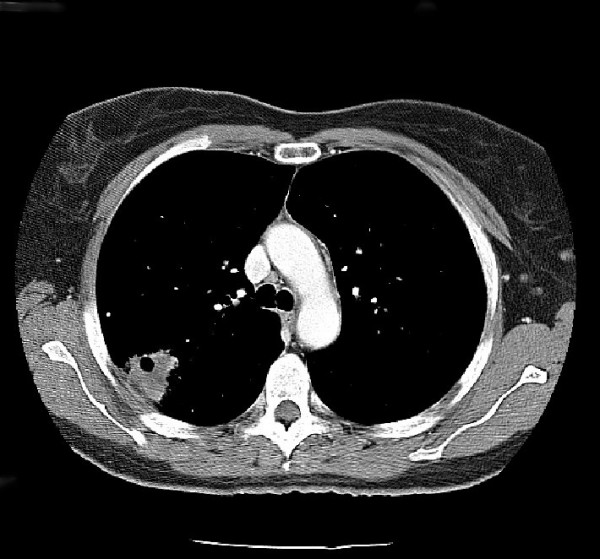}
\end{wrapfigure}

\textbf{Caption:}\\
{\footnotesize
Figure 4: Contrast-enhanced CT scan of the chest for patient \#1. A large cavitating lesion was identified in the right upper lobe with pleural involvement and likely extension into the chest wall. In addition, right hilar lymphadenopathy was suspected. This was the primary malignant lesion in this patient.\vspace{5pt}
}

{\bf Sentences with in-line figure mentions}:

\begin{itemize}[noitemsep,topsep=0pt,leftmargin=10pt]
\item[-] {\footnotesize Computerized tomography ( CT ) scans of the chest ( Figure 4 ) were obtained.}
\item[-] {\footnotesize ( Figure 4 ) and demonstrated a large cavitating lesion in the posterior aspect of the right upper lobe with probable pleural involvement likely representing the primary lesion.}
\end{itemize}

\tcbline

{\footnotesize\bf Conversation:}\\
\parbox{1\textwidth}{\scriptsize
\begin{alltt}
User: Where is the lesion located?\\

Assistant: The lesion is located in the posterior aspect of the right upper lobe of the lung.\\

User: Is it extending into the chest wall?\\

Assistant: Yes, the CT scan suggests that the large cavitating lesion in the right upper lobe is likely extending into the chest wall.\\

User: Can you explain more about the cavitating lesion?\\

Assistant: Certainly. A cavitating lesion refers to an abnormal mass or growth within the lung tissue that has a cavity or hollow space inside it. This cavity can be formed due to the necrosis (death) of the tissue inside the lesion, leading to the breakdown and disintegration of the affected tissue. Cavitating lesions can be caused by various factors, such as infections (like tuberculosis or fungal infections), autoimmune diseases, or malignancies (cancer). In the context of the CT scan provided, the large cavitating lesion in the right upper lobe of the lung is suggestive of a malignant (cancerous) growth. The cavity within the lesion may be a result of the rapid and uncontrolled growth of cancer cells, which can lead to the destruction of surrounding healthy tissue. It's important to note that further diagnostic tests, such as a biopsy, would be required to confirm the malignancy and determine the specific type of cancer involved.
\end{alltt}}
\end{AIbox}
\caption{
An instance of our GPT-4 generated instruction-following data. 
Top: The figure and caption were extracted from a PubMed Central full-text article ~\cite{pringle2006appropriate}, along with the corresponding citances (mentions of the given figure in the article). Bottom: The instruction-following data generated by GPT-4 using the text only (caption and citances). Note that the image is not used to prompt GPT-4; we only show it here as a reference.
%One example to illustrate the instruction-following data. The top block shows the caption, sentences with in-line figure mentions, and the medical image (all collected from \cite{pringle2006appropriate}) The bottom block shows instruction-following data regarding the medical image. Note that the image is not used to prompt GPT-4; we only show it here as a reference.
} \label{fig:instruct-ex}
\vspace{-3mm}
\end{figure}

\vspace{-2mm}
\section{Biomedical Visual Instruction-Following Data}
\label{sec:visual_instruc_data}
\vspace{-3mm}

There are a lack of multimodal biomedical datasets to train an instruction-following assistant. To fill this gap, we create the first dataset of its kind from widely existing biomedical image-text pairs, through a machine-human co-curation procedure. It consists of two sets, concept alignment and instruction-following, which are used at different training stages, described in \Cref{sec:training}.

\paragraph{Biomedical Concept Alignment Data.}
For a biomedical image $\Xmat_{\texttt{v}}$ and its associated caption $\Xmat_{\texttt{c}}$, we sample a question $\Xmat_{\texttt{q}}$, which asks to describe the biomedical image.
With $(\Xmat_{\texttt{v}},\Xmat_{\texttt{c}},\Xmat_{\texttt{q}})$, we create a single-round instruction-following example:
\begin{center}
$
\texttt{Human}: \Xmat_{\texttt{q}} ~\Xmat_{\texttt{v}}     \texttt{<STOP>} \backslash \texttt{n}~
    \texttt{Assistant}: 
    \Xmat_{\texttt{c} } 
     \texttt{<STOP>} \backslash \texttt{n}
$ 
\end{center}
Depending on the length of caption, the question that is sampled either asks to describe the image \emph{concisely} or \emph{in detail}.
Two lists of questions are provided in \Cref{sec:data}.
In practice, 25\% of captions have length less than 30 words in PMC-15M~\cite{zhang2023large}, and thus 30 words is used as the cutoff point to determine which list to choose.
We sample 600K image-text pairs from PMC-15M. Though this dataset only presents one-single task instructions, \ie image captioning, it contains a diverse and representative set of biomedical concept samples from the original PMC-15M~\cite{zhang2023large}.

\paragraph{Biomedical Instruction-Tuning Data.}
To align the model to follow a variety of instructions, we present and curate diverse instruction-following data with multi-round conversations about the provided biomedical images, by prompting language-only GPT-4.
Specifically, given an image caption, we design instructions in a prompt that asks GPT-4 to generate multi-round questions and answers in a tone as if it could see the image (even though it only has access to the text).
Sometimes the image caption is too short for GPT-4 to generate meaningful questions and answers.
To provide more context regarding the image, we also create a prompt that includes not only captions but also sentences from the original PubMed paper that mentions the image.
We also manually curate few-shot examples in the prompt to demonstrate how to generate high-quality conversations based on the provided caption and context.
See \Cref{sec:prompts} for the prompt and few-shot examples.
To collect image captions and their context, we filter PMC-15M to retain the images that only contain a single plot.
From them, we sample 60K image-text pairs from the five most common imaging modalities: CXR (chest X-ray), CT (computed tomography),  MRI (magnetic resonance imaging), histopathology, and gross (\ie macroscopic) pathology.
We then extract sentences that mention the image from the original PubMed paper as additional context to the caption, inspired by the observations that external knowledge helps generalization~\cite{lewis2020retrieval,liu2023learning}.

\begin{figure*}[t!]%\vspace{-25pt}
	\vspace{-0mm}\centering
	\begin{tabular}{c c}
		\hspace{-3mm}
\includegraphics[height=6.0cm]{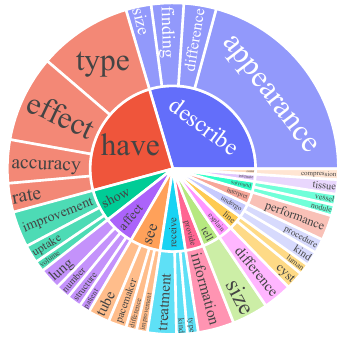} & 
\includegraphics[height=6.0cm]{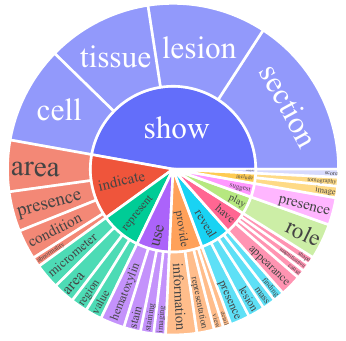}  \\
		(a) Instruction
		&
		(b) Responses \vspace{-0mm} \\
		\hspace{-3mm}

  \includegraphics[height=2.8cm]{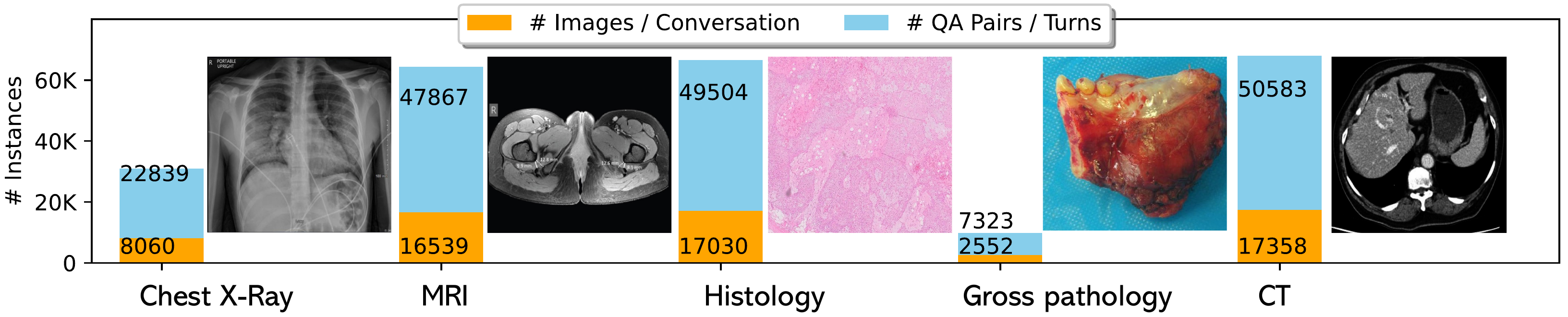}
  \hspace{-65mm}
  &  \\
  \hspace{-3mm}
  % \multicolumn{4}{c}{}
  (c) Frequencies of images and QA pairs on the five domains.
  \hspace{-65mm}
  & 
   \\
		% \hspace{-3mm}
  % % \multicolumn{4}{c}{}
  % \includegraphics[height=3.5cm]{figures/cmp_seq_len_freq_gpt3_gpt4.pdf}
  % \hspace{-65mm}
  % &  \\
  % \hspace{-3mm}
  % % \multicolumn{4}{c}{}
  % (d) Frequencies of output sequence lengths
  % \hspace{-65mm}
  % & 
	\end{tabular}
	\vspace{-2mm}
	\caption{The data statistics of biomedical multimodal instruction-following data: (a,b) The root verb-noun pairs of instruction and responses, where the inner circle of the plot represents the root verb of the output response, and the outer circle represents the direct nouns.
    (c) The distribution of images and QA pairs on the five domains, one image is shown per domain.
    The domain example images are from \cite{ayoub2021covid,papavasiliou2021quadratus,basak2021chondromyxoid,mirmohammad2021conventional,zafar2021delayed}. 
	 }
	\label{fig:data_stats}
 \vspace{2mm}
\end{figure*}

An example of instruction-following data is shown in \Cref{fig:instruct-ex} shows, and the data statistics is shown Figure~\ref{fig:data_stats}. We have produced three versions of instruct data when iteratively improving the data quality:  
$(i)$ {\it 60K-IM}. The aforemenioned dataset that considers inline mentions (IM) as the context.
$(ii)$ {\it 60K}. A dataset of similar size (60K samples) without IM in self-instruct generation.
$(iii)$ {\it 10K}. A smaller dataset (10 samples) without IM. They are used to ablate our data generation strategies and their impact on trained \shortname{} in experiments.

\vspace{-2mm}
\section{Adapting Multimodal Conversational Models to the Biomedical Domain}
\label{sec:training}
\begin{figure}[t!]
\centering  
\vspace{-4mm}
\includegraphics[height=2.7cm]{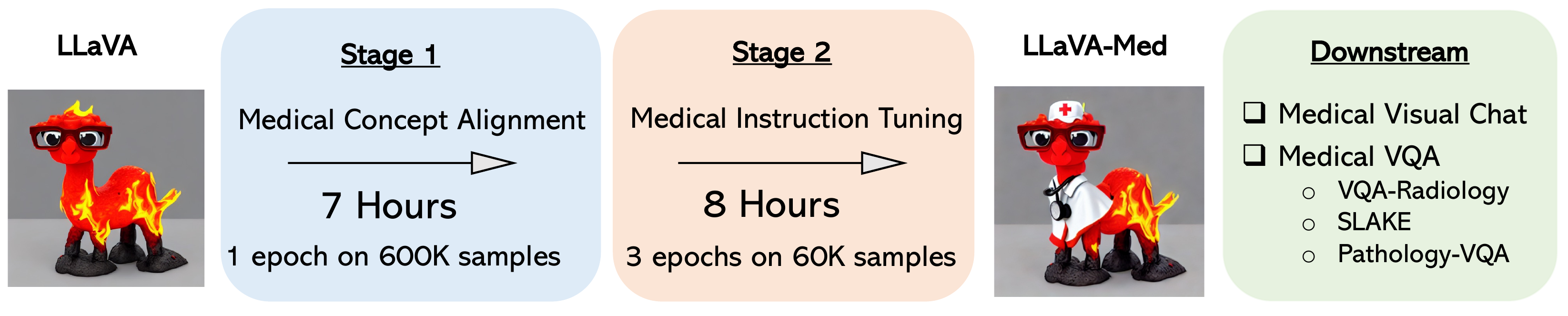}
\vspace{-1mm}
\caption{\shortname{} was initialized with the general-domain LLaVA and then continuously trained in a curriculum learning fashion (first biomedical concept alignment then full-blown instruction-tuning). We evaluated \shortname{} on standard visual conversation and question answering tasks.}
\label{fig:llava_med_training}  
  \vspace{-3mm}
\end{figure}

\vspace{-2mm}
We employ LLaVA,a general-domain multimodal conversation model~\cite{liu2023visual}, as the initial general-domain \ac{LM}, and continuously train the model to the biomedical domain. The same network architecture is utilized, where a linear projection layer connects the vision encoder and the language model. 
For \shortname{} model training, we use a two-stage procedure, illustrated in Figure~\ref{fig:llava_med_training}. 

\paragraph{Stage 1: Biomedical Concept Feature Alignment.} To balance between concept coverage and training efficiency, we filter PMC-15M to 600K image-text pairs. These pairs are converted to instruction-following data using a naive expansion method: instructions simply presents the task of describing the image. For each sample, given the language instruction and image input, we ask the model to predict the original caption. In training, we keep both the visual encoder and \ac{LM} weights frozen, and only update the projection matrix.  In this way, the image features of vast novel biomedical visual concepts can be aligned to their textual word embeddings in the pre-trained \ac{LM}. This stage can be understood as expanding the vocabulary of aligned image-text tokens to the biomedical domain.
  
\paragraph{Stage 2: End-to-End Instruction-Tuning.} 
%
%For specific use scenarios, 
We only keep the visual encoder weights frozen, and continue to update both the pre-trained weights of the projection layer and \ac{LM}. To train the model to follow various instructions and complete tasks in a conversational manner, we develop a biomedical chatbot by fine-tuning our model on the biomedical language-image instruction-following data collected in Section~\ref{sec:visual_instruc_data}. As demonstrated in the experiments to be described later, the \shortname{} model at this stage is able to not only be served as a biomedical visual assistant to interact with users, but also achieve good zero-shot task transfer performance when evaluated on well-established biomedical \ac{VQA} datasets.

\paragraph{Fine-tuning to Downstream Datasets.} 
For some specific biomedical scenarios, there is a need of developing highly accurate and dataset-specific models to improve the service quality of the assistant.
We fine-tune \shortname{} after the two-stage training on three biomedical \ac{VQA} datasets~\cite{lu2022learn}, covering varied dataset sizes and diverse biomedical subjects. Given a biomedical image as context, multiple natural language questions are provided, the assistant responds in free-form text for both the close-set and open-set questions, with a list of candidate answers constructed in the prompt for each close-set question.

\paragraph{Discussion.} We discuss three favorable properties/implications of \shortname{}:
$(i)$ {\it Affordable development cost.}
Instead of scaling up data/model for the best performance, we aim to provide affordable and reasonable solutions with low development cost: it takes 7 and 8 hours for stage 1 and 2 on 8 40G A100 GPUs, respectively (see Table~\ref{tab:running_time} for detailed numbers). 
$(ii)$ {\it A recipe for many domains.}
Though this paper focuses on biomedical domains, the proposed adaptation procedure is generalizable to other vertical domains such as gaming and education, where novel concepts and domain knowledge are needed to build a helpful assistant. Similar to the {\it don't stop pre-training} argument in~\cite{gururangan2020don}, we consider a scalable pipeline to create domain-specific instruct data from large unlabelled data, and advocate {\it don't stop instruction-tuning} to build customized LMM. 
$(iii)$ {\it Low serving cost.}
While the model size of general LMM can be giant and serving cost can be prohibitively high, customized LMM has its unique advantages in low serving cost.
$(iv)$ {\it Smooth Model Adaptation.}
Alternatively, the network architecture allows us to initialize the vision encoder from BioMedCLIP~\cite{zhang2023large}, or initialize the language model from Vicuna~\cite{vicuna}, which may lead to higher performance. However, adapting from LLaVA smooth adaptation as a chatbot, where model's behaviors transit from layperson to a professional assistant that is able to provide helpful domain-specific response.

% \chunyl{Provide detailed cost to show the efficiency of the adaptation process.}

% \chunyl{Provide a figure to illustrate the adaptation process.}

\vspace{-2mm}
\section{Experiments}
\label{sec:experiments}
\vspace{-2mm}
We conduct experiments to study two key components, the quality of the produced multimodal biomedical instruction-following data, and performance of \shortname{}.
We consider two research evaluation settings: (1) What is the performance of \shortname{} as an open-ended biomedcal visual chatbot? (2) How does \shortname{} compare to existing methods on standard benchmarks? To clarify, throughout the entire experiments, we only utilize the language-only GPT-4.

\subsection{Biomedical Visual Chatbot}
\label{sec:exp_chat}

% \chunyl{Quantitative and qualitative comparisons of the instruction-following performance on (1) three models: \shortname{}, LLaVA, GPT4; (2) Ablation studies of three versions of instruct data  .}

To evaluate the performance of \shortname{} on biomedical multimodal conversation, we construct an evaluation dataset with 193 novel questions.
For this test dataset, we randomly selected 50 {\it unseen} image and caption pairs from PMC-15M, and generate two types of questions: conversation and detailed description. The conversation data is collected using the same self-instruct data generation pipeline as for the 2nd stage. Detailed description questions were randomly selected from a fixed set \cite{liu2023visual} of questions to elicit detailed description responses.

We leverage GPT-4 to quantify the correctness of the model answer to a question when given the image context and caption. GPT-4 makes a reference prediction, setting the upper bound answer for the teacher model.  We then generate response to the same question from another LMM. Given responses from the two assistants (the candidate LMM and GPT-4), the question, figure caption, and figure context, we ask GPT-4 to score the helpfulness, relevance, accuracy, and level of details of the responses from the two assistants, and give an overall score on a scale of 1 to 10, where a higher score indicates better overall performance. GPT-4 is also asked to provide a comprehensive explanation the evaluation, for us to better understand the models. We then compute the relative score using GPT-4 reference score for normalization.

\begin{table}[ht!]
\centering
\tablestyle{4pt}{1.2}
% \scalebox{0.85}{
\begin{tabular}{l|cc|ccccc|c}  
% \toprule
&  \multicolumn{2}{c|}{\textbf{Question Types}}  & \multicolumn{5}{c|}{\bf Domains}  & {\bf Overall}    \\
&   Conversation &  Description & CXR & MRI & Histology & Gross & CT &   \\
(Question Count)  & (143) & (50) & (37) & (38) & (44) & (34) & (40) & (193)  \\ \shline
 LLaVA & 39.4 & 26.2 & 41.6 & 33.4 & 38.4 & 32.9 & 33.4 & 36.1 \\ \hline
 \shortname{} & & &  & & & & \\
  \qquad Stage 1 & 22.6 & 25.2 & 25.8 & 19.0 & 24.8 & 24.7 & 22.2 & 23.3 \\
  \qquad 10K & 42.4 & 32.5 & 46.1 & 36.7 & 43.5 & 34.7 & 37.5 & 39.9  \\
  \qquad 60K & 53.7 & 36.9 & 57.3 & 39.8 & 49.8 & 47.4 & 52.4 & 49.4  \\
  \qquad 60K-IM  & 55.1 & 36.4 & 56.2 & 40.4 & 52.7 & 51.8 & 50.1 & 50.2  \\ 
% \bottomrule
\end{tabular}  
% }
\vspace{1mm}
\caption{Performance comparison of mulitmodal chat instruction-following abilities, measured by the relative score via language GPT-4 evaluation.}
\label{tab:chat_perf}
\end{table}

% \begin{table}[ht!]
% \centering
% \tablestyle{5pt}{1.2}
% % \scalebox{0.85}{
% \begin{tabular}{c p{1.2cm}|cc|ccccc|c}  
% % \toprule
% & &  \multicolumn{2}{c|}{\textbf{Question Types}}  & \multicolumn{5}{c|}{\bf Domains}  & \multirow{2}{*}{\bf Overall}    \\
% &  &  Conversation &  Description & CXR & MRI & Histology & Gross & CT Scan &   \\
% \multicolumn{2}{l|}{Question Count}  & 143 & 50 & 37 & 38 & 44 & 34 & 40 & 193  \\ \shline
%  \multicolumn{2}{c|}{LLaVA} & 39.4 & 26.2 & 41.6 & 33.4 & 38.4 & 32.9 & 33.4 & 36.1 \\ %\vspace{1mm} \\ 
% \multirow{4}{*}{\rotatebox{90}{\footnotesize \shortname{} \hspace{-1mm}} } 
% &  Stage 1 & 22.6 & 25.2 & 25.8 & 19.0 & 24.8 & 24.7 & 22.2 & 23.3 \\
% &  10K & 42.4 & 32.5 & 46.1 & 36.7 & 43.5 & 34.7 & 37.5 & 39.9  \\
% &  60K & 53.7 & 36.9 & 57.3 & 39.8 & 49.8 & 47.4 & 52.4 & 49.4  \\
% &  60K-IM  & 55.1 & 36.4 & 56.2 & 40.4 & 52.7 & 51.8 & 50.1 & 50.2  \\
% % \bottomrule
% \end{tabular}  
% % }
% \vspace{1mm}
% \caption{Performance comparison of mulitmodal chat instruction-following abilities, measured by the relative score via GPT-4 evaluation.}
% \label{tab:chat_perf}
% \end{table}

The results are reported in Table~\ref{tab:chat_perf}. 
\shortname{} with Stage-1 training alone is insufficient as a chatbot, as it loses its ability to follow diverse instructions, though biomedical concept coverage is improved. 
\shortname{} with the full two-stage training consistently outperforms the general domain LLaVA, and training with larger instruct data (from 10K to 60K samples) leads to higher performance. When inline mentions are considered in self-instruct, the generated data 60K-IM slightly improves the chat ability. The results demonstrate the effectiveness of the strategies in biomedical instruction-following data collection as well as the value of dataset assets.
Overall, for the best \shortname{}, it matches the 50.2\% performance of GPT-4. Note that GPT-4 generates response by considering ground-truth caption and golden inline mentions, without understanding the images. Though not a fair comparison between LMMs and GPT-4, GPT-4 is a consistent and reliable evaluation tool.

\begin{table}
  \begin{minipage}{0.99\textwidth}
\centering  
\small
\vspace{-4mm}
\scalebox{0.83}{
\begin{tabular}{p{2.1cm} p{13.2cm}  }
\toprule
 \multicolumn{2}{l}{\bf Visual input example, Biomedical image from  \cite{bharat2020vaping}}  \\
\midrule
% Caption & Chest x-ray (CXR) on Day 2 of admission post-intubation (yellow line showing the level of the endotracheal tube). Rapidly worsening ground-glass opacities are seen throughout the lungs with relative subpleural sparing (red arrows). \cite{bharat2020vaping}
% &  \includegraphics[height=3.5cm]{figures/example_imgs/visual_chat_example.jpg} \\
&  \includegraphics[height=3.5cm]{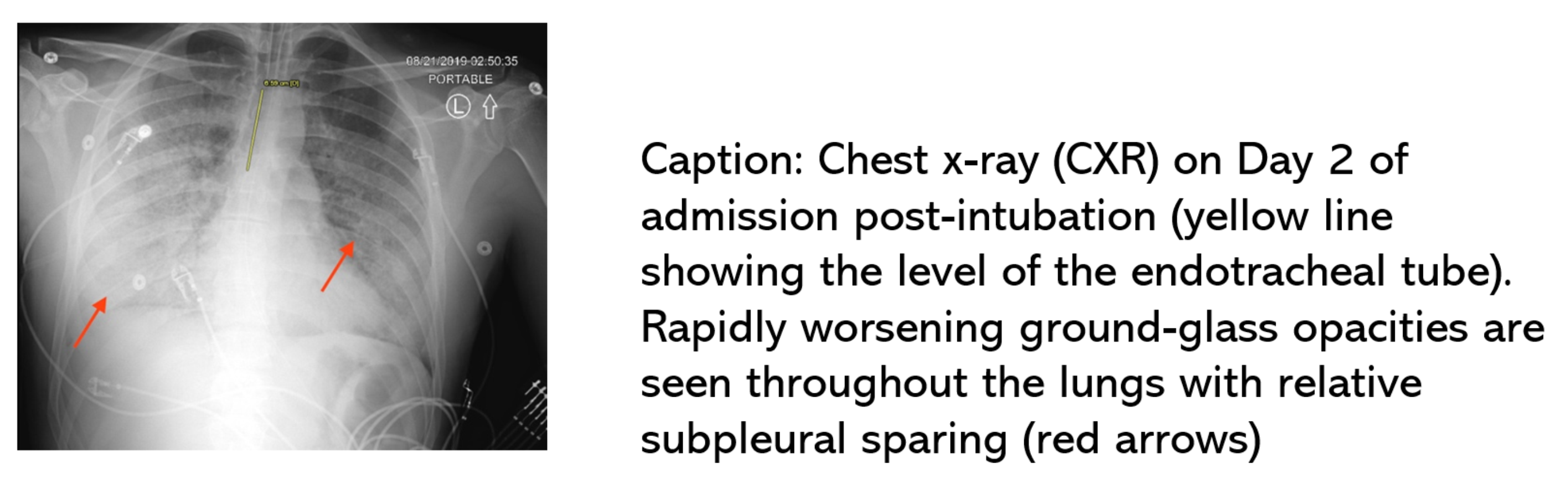} \\
 & {\tiny Source:~\url{https://www.ncbi.nlm.nih.gov/pmc/articles/PMC7398722/}} \\
User & What is shown in this image?\\
\shortname{} & The image is a chest X-ray (CXR) that shows bilateral patchy infiltrates, which are areas of increased opacity in the lungs. These infiltrates can be indicative of various lung conditions, such as infections, inflammation, or other lung diseases.
\\
User & Are there any devices implanted in this patient? \\
\shortname{} & Yes, it appears that the patient has several devices implanted. These devices may include endotracheal tubes, central venous catheters, and/or pulmonary artery catheters. These devices are typically used for monitoring and managing various medical conditions, as well as for providing support during surgical procedures.
\\
\midrule
User & What is shown in this image?  \\
LLaVA &  The image shows a ribcage with wires coming out of it, indicating that it is likely a medical illustration or diagram.  \\
User &  Are there any devices implanted in this patient?  \\
LLaVA  &  Yes, there are several devices protruding from the patient's chest, which are likely medical devices used for treatment or monitoring.  \\
\midrule
\textcolor{gray}{User} & \textcolor{gray}{What is shown in this image?}  \\
\textcolor{gray}{Language GPT-4} &  \textcolor{gray}{The image shows a chest x-ray taken on Day 2 of a patient's admission after intubation. You can see the endotracheal tube marked by a yellow line, and rapidly worsening ground-glass opacities throughout the lungs. There is also relative subpleural sparing, which is indicated by red arrows.}  \\
\textcolor{gray}{User} &  \textcolor{gray}{Are there any devices implanted in this patient?}  \\
\textcolor{gray}{Language GPT-4}  & \textcolor{gray}{Yes, there is an endotracheal tube implanted in the patient, as indicated by the yellow line in the image.}  \\
\bottomrule
\end{tabular}
}
\vspace{1mm}
\captionof{table}{Example comparison of medical visual chat and reasoning capabilities. The language-only GPT-4 is considered as the performance upper bound, as the golden captions and inline mentions are fed into GPT-4 as the context, without requiring the model to understand the raw image.}
% briefly say what to notice.
% The prompt requires image understanding.}  

\vspace{-5mm}
\label{tab:visual_chat_example}  
  \end{minipage}
\end{table}

In Table~\ref{tab:visual_chat_example}, we provide examples on the biomed visual conversations of different chatbots. \shortname{} precisely answers the questions with biomedical knowledge, while LLaVA behaves like a layperson, who hallucinate based on commonsense. Since the multimodal GPT-4 is not publicly available, we resort to language-only GPT-4 for comparison. We feed golden captions and inline mentions into GPT-4 as the context, it generates knowledgeable response through re-organizing the information in the conversational manner.

\subsection{Performance on Established Benchmarks}
\label{sec:exp_benchmarks}
\paragraph{Dataset Description.} We train and evaluate \shortname{} on three biomedical \ac{VQA} datasets. The detailed data statistics are summarized in Table~\ref{tab:data_stat_three_datasets}.
% The characteristics of these datasets are listed in Table 1, and sample images and their corresponding question–answer pairs from the training set are shown in Figures 5 and 6.

\begin{table}[!ht]
\centering 
\tablestyle{4pt}{1.2}
\begin{tabular}{l|cc|ccc|ccc}  
  & \multicolumn{2}{c|}{\bf VQA-RAD} & \multicolumn{3}{c|}{\bf SLAKE} & \multicolumn{3}{c}{\bf PathVQA}  \\
Dataset   & Train    & Test   & Train  & Val  & Test    & Train  & Val & Test  \\ 
     \shline
\# Images & 313  &  203 &   450  & 96 &  96 & 2599 & 858 & 858 \\
 \# QA Pairs & 1797   & 451   & 4919  &  1053    &  1061 & 19,755   & 6279   & 6761   \\
 \# Open &   770  &  179  &   2976  &  631   &    645   &   9949  &   3144 &  3370 \\
 \# Closed &   1027 &  272 &     1943 &  422  &     416  &    9806 &    3135 &  3391    
\end{tabular}  
\caption{Dataset statistics. For SLAKE, only the English subset is considered for head-to-head comparison with existing methods.}
\label{tab:data_stat_three_datasets}  
\end{table}

\begin{table}[t!]  
\centering  
\begin{subtable}{1.0\textwidth} 
\centering 
\tablestyle{3pt}{1.2}
\begin{tabular}{ll|ccc|ccc|ccc}  
 & & \multicolumn{3}{c|}{\bf VQA-RAD} & \multicolumn{3}{c|}{\bf SLAKE} & \multicolumn{3}{c}{\bf PathVQA}  \\
Method   &  & Ref   & Open   & Closed   & Ref   & Open   & Closed    & Ref  & Open &  Closed  \\ \shline
\multicolumn{11}{l}{\it Supervised finet-tuning results with our own experiment runs } \\
 % \rowcolor{Gray}   
% \multirow{2}{*}{Zero-shot} 
 % \multicolumn{2}{l|}{ LLaVA } 
 % &   & 16.98 & 66.49 &   & 26.82 & 66.91 &   & 8.74 & 57.80 \\
 % \rowcolor{Gray}  
 % \multicolumn{2}{l|}{  \shortname{}  (10K) } 
 % & & 24.84 & 76.84 & &  31.50 & 71.70 & &  8.49 & \textcolor{blue}{\bf 89.20} \\
 % \rowcolor{Gray}
 % \multicolumn{2}{l|}{  \shortname{}  (60K) }
 % &   & 31.77 &  \textcolor{blue}{\bf 86.81}  &   & 37.48 & 74.29 &   & 11.46 & 83.73 \\
 %  \rowcolor{Gray}
 % \multicolumn{2}{l|}{  \shortname{}  (60K-IM) }
 % &  &  32.21 & \textcolor{blue}{\bf 86.39} &  &  39.17 & 79.50 &  &  12.30 & 85.30 \\
 % \rowcolor{Gray}
 % Fine-tuned % \multirow{2}{*}{Fine-tuned} 
 % &  LLaVA  &  & 16.84 & 89.02 & & 69.03 & 73.25 & & 7.74 & 72.37 \\
 % \rowcolor{Gray}
 % & \shortname{}  & & 26.41 & {\bf 89.89} & & 79.33 & 79.32 &  & 17.18 & {\bf 99.42}  \\
\multicolumn{2}{l|}{LLaVA} &   & 50.00 & 65.07 & & 78.18 & 63.22 & & 7.74 & 63.20  \\ 
   \rowcolor{Gray}
\multicolumn{2}{l|}{\shortname{} (From LLaVA)} &   & 61.52 & {\bf 84.19} & & 83.08 & 85.34  & &  37.95   & {\bf 91.21}\\  
 \rowcolor{Gray}
\multicolumn{2}{l|}{\shortname{} (From Vicuna)} &   & 64.39 & 81.98 & & {\bf 84.71} & 83.17  & &  38.87  & {\bf 91.65}\\ 
   \rowcolor{Gray}
\multicolumn{2}{l|}{\shortname{} (BioMed CLIP)} &   & 64.75 & 83.09 & & {\bf 87.11} & 86.78 & & 39.60 & {\bf 91.09}\\ 

 \hline
  \multicolumn{11}{l}{\it Representative \& SoTA methods with numbers reported in the literature } \\  
\multicolumn{2}{l|}{VL Encoder–Decoder~\cite{bazi2023vision}} &  71.49 & & 82.47 &  & &  & 71.49 & & 85.61\\
\multicolumn{2}{l|}{Q2ATransformer~\cite{liu2023q2atransformer}} &  79.19 & & 81.20 & & &  & 54.85 & & 88.85 \\
\multicolumn{2}{l|}{Prefix T. Medical LM~\cite{van2023open}} &  & & & 84.30 & & 82.01 & 40.00 & & 87.00\\ 
\multicolumn{2}{l|}{PubMedCLIP~\cite{eslami2023pubmedclip}} &  60.10 & & 80.00 & 78.40 & & 82.50 & & & \\
\multicolumn{2}{l|}{BiomedCLIP~\cite{zhang2023large}} &  67.60 & & 79.80 & 82.05 & & 89.70 & & & \\
\multicolumn{2}{l|}{M2I2~\cite{li2022self}} & 66.50 & & 83.50 & 74.70 & & 91.10 & 36.30 & & 88.00 
\end{tabular}  
\caption{Comparison with prior state-of-the-art supervised methods. For open-ended questions, prior methods still formulate the problem as classification among distinct answers in the training set, which may overestimate their generalizability as these datasets are unusual in that the test answers are almost always present in training.
%We quote the open-set performance of existing methods in column {\it Ref}, which selects the best answer from a pre-defined candidate set.
}  
  \end{subtable}

\begin{subtable}{1.0\textwidth} 
\centering 
\tablestyle{3pt}{1.2}
\begin{tabular}{lccc|rr|rr|rr|c}  
\multicolumn{4}{c|}{\bf \shortname{} Model Variants}    & \multicolumn{2}{c|}{\bf VQA-RAD} & \multicolumn{2}{c|}{\bf SLAKE} & \multicolumn{2}{c|}{\bf PathVQA} & \multirow{2}{*}{Average}  \\
Instruct &   Stage 1 & Stage 2  & FT  & Open    & Closed   & Open    & Closed    & Open  &  Closed & \\ 
     \shline
 \multicolumn{11}{l}{\it CLIP Vision Encoder~\cite{radford2021learning}, 7B Language Model} \\  
0 & 1 & 0 & 0 & 15.27 & 12.50 & 18.55 & 13.46 & 6.26 & 13.51 & 13.26  \\
 0   & 3& 0 & 0  &15.33 & 15.44 & 23.61 & 15.38 & 6.35 & 14.74 & 15.14  \\
 10K & 1& 3 & 0  & 25.79 & 57.35 & 31.50 & 51.68 & 8.49 & 59.66 & 39.08 \\
 10K & 3& 3 & 0  & 28.44 & 59.56 & 22.63 & 43.99 & 5.40 & 52.67 & 35.45 \\ 
 10K & 1& 3 & 1  & 36.39 & 55.88 & 71.64 & 56.49 & 25.50 & 82.87 & 54.79 \\
 10K & 1& 3 & 3  & 18.59 & 55.51 & 78.60 & 63.46 & 34.02 & 86.94 & 56.19  \\ 
 60K & 1& 1 & 0  & 29.80 & 55.15 & 38.08 & 50.00 & 11.70 & 59.66 & 40.73  \\

 60K & 1& 3 & 0  & 29.67 & 60.29 & 35.53 & 53.85 & 11.76 & 53.20 & 40.72  \\
%  60K & 1& 1 & 1  & 26.52 & 52.21 & 70.94 & 57.21 & 24.83 & 81.75 & 52.24  \\
%  60K & 1& 1 & 3  & 51.66 & 62.87 & 28.93 & 25.72 & 32.39 & 87.26 & 48.14 \\
 60K & 1& 3 & 1  & 22.63 & 58.09 & 72.75 & 54.33 & 24.19 & 71.60 & 50.60 \\

 60K & 1& 3 & 3  & 54.12 & 64.71 & 79.33 & 64.90 & 17.18 & 71.37 & 58.60  \\
60K-IM & 1& 1 & 0  & 29.67 & 61.40 & 38.44 & 52.40 & 11.41 & 56.24 & 41.59 \\ 
60K-IM & 1& 3 & 0  & 28.23 & 61.40 & 39.17 & 52.16 & 12.30 & 54.05 & 41.22 \\
% 60K-IM & 1& 9 & 0  &27.94 & 60.66 & 34.59 & 55.77 & 10.33 & 52.67 & 40.33 \\
60K-IM & 1& 3 & 1  & 28.61 & 56.25 & 70.58 & 54.57 & 11.17 & 59.19 & 46.73 \\
60K-IM & 1& 3 & 3  & 55.50 & 66.54 & 80.57 & 64.18 & 35.88 & 89.15 & 65.30 \\
60K-IM & 1& 3 & 9  & 66.26 & 80.88 & 82.30 & 84.86 & 37.59 & 91.54 & 73.90 \\
  \rowcolor{Gray}
60K-IM & 1& 3 & 15  & 61.53 & 84.19 & 83.08 & 85.34 & 37.95 & 91.21 & 73.88  \\
60K-IM & 1& 3 & 18  & 61.37 & 81.25 & 84.24 & 83.17 & 37.88 & 91.39 & 73.22  \\
 \multicolumn{11}{l}{\it CLIP Vision Encoder~\cite{radford2021learning}, 13B Language Model} \\  
 60K-IM & 1& 3 & 0  & 31.66 & 61.40 & 37.71 & 49.76 & 11.34 & 49.63 & 40.25  \\
 60K-IM & 1& 3 & 9  & 64.58 & 77.94 & 84.97 & 85.58 & 38.82 & 92.39 & 74.05   \\
 \multicolumn{11}{l}{\it BioMed CLIP Vision Encoder~\cite{zhang2023large}, 7B Language Model } \\  
 60K-IM & 1& 3 & 0  & 37.84 & 60.66 & 39.73 & 54.33 & 11.65 & 49.07 & 42.21 \\
   \rowcolor{Gray}
 60K-IM & 1& 3 & 9  & 64.75 & 83.09 & 87.11 & 86.78 & 39.60 & 91.09 & 75.40  \\
\hline
LLaVA  & 0 & 0 & 0   & 20.74 & 59.19 & 26.82 & 50.24 & 8.74 & 45.65 & 35.23 \\
% LLaVA  & 0 & 0 & 3  & 50.00 & 65.07 & 78.18 & 63.22 & 7.74 & 63.20 & 54.57  \\ 
\end{tabular}  
\caption{Ablation studies with varying number of training epochs at different stages. ``FT'' is Fine-Tuning. 60K-IM indicates the instruct data generated with inline mentions.
The gray rows are zero-shot performance of \shortname{} trained with different instruct data, they are selected to show in subtable (a).}  
  \end{subtable}  

\caption{Quantitative results on three established biomedical \ac{VQA} datasets. For open-set questions, we report the recall for our free-form text generation method in column {\it Open}. For closed-set questions, we report the accuracy in column {\it Closed}. Bold indicates \shortname{} achieves new SoTA.}  
\label{tab:medical_vqa_model_performance}  
\end{table}

\begin{table}[!ht]
\centering 
\tablestyle{4pt}{1.2}
\begin{tabular}{cc|ccc|cc|cc|cc}  
 \multicolumn{2}{c|}{\bf Stage 1}   &  \multicolumn{3}{c|}{\bf Stage 2}   & \multicolumn{2}{c|}{\bf VQA-RAD} & \multicolumn{2}{c|}{\bf SLAKE} & \multicolumn{2}{c}{\bf PathVQA}  \\
 1  &  3  & Instruct & 1 & 3  & 1    & 3   & 1    & 3    & 1  &  3  \\ 
     \shline
\multirow{2}{*}{6.8}  & \multirow{2}{*}{19.4}  
& 10K & 0.6 & 1.8  & 
\multirow{2}{*}{0.3}  &  \multirow{2}{*}{0.6}  & 
\multirow{2}{*}{0.6}  &  \multirow{2}{*}{1.0}  & 
\multirow{2}{*}{1.0}  &  \multirow{2}{*}{2.5}  \\
   &   & 60K & 2.6 & 8.0  &  &  & & & &      
\end{tabular}  
\caption{Running time (hours) for 1 and 3-epoch training at each stage, with batch size 128 on eight A100 GPUs.}
\label{tab:running_time}  
\end{table}

\begin{itemize}[leftmargin=5.5mm]
\setlength{\itemsep}{3pt}
\item 
{\it VQA‐RAD}~\cite{lau2018dataset} contains 3515 QA pairs generated by clinicians and 315 radiology images that are evenly distributed over the head, chest, and abdomen. Each image is associated with multiple questions. Questions are categorized into 11 categories: abnormality, attribute, modality, organ system, color, counting, object/condition presence, size, plane, positional reasoning, and other. Half of the answers are closed‐ended (\ie yes/no type), while the rest are open‐
ended with either one‐word or short phrase answers.

\item
{\it SLAKE}~\cite{liu2021slake} is a Semantically-Labeled Knowledge-Enhanced dataset for medical \ac{VQA}. It consists of 642 radiology images and over 7000 diverse QA pairs annotated by experienced physicians, where the questions may involve external medical knowledge (solved by provided medical knowledge graph), and the images are associated with rich visual annotations, including  semantic segmentation masks and object detection bounding boxes.
Besides, SLAKE includes richer modalities and covers more human body parts than the currently available dataset, including brain, neck, chest, abdomen, and pelvic cavity. Note SLAKE is bilingual dataset with English and Chinese. When compared with existing methods, we only consider the English subset.

\item
{\it PathVQA}~\cite{he2020pathvqa} is a dataset of pathology images. It contains a total of 4998
pathology images with 32,799 QA pairs. Every image has several questions that relate to multiple aspects such as location, shape, color, appearance, etc. The questions are categorized into two types,
with several varieties: open-ended questions such as why, what, how, where, \etc, and closed-ended questions.

\end{itemize}

\paragraph{Evaluation Metrics.} For the closed-set questions, we report the accuracy. For open-set questions, we use recall to evaluate the ratio that ground-truth tokens appear in the generated sequences. In the literature, the unique answers in the training set are considered as the answer candidates, from which the models can select to predict answers for testing questions. Since we do not provide any constraint for the responses to open-set questions, our formulation is closer to open-set nature, but is intrinsically harder.

\paragraph{Comparisons with SoTA.} We compare \shortname{} with the general domain LLaVA and existing representative methods in Table~\ref{tab:medical_vqa_model_performance} (a). 
First, All \shortname{} variants outperform LLaVA. While the difference of language model initialization from LLaVA or Vicuna is minor, the initialization of vision encoder from BioMed CLIP is slightly better than from general-domain CLIP.
Second, the fine-tuning performance of \shortname{} is higher than supervised SoTA on the closed-set questions on VQA-RAD and PathVQA. This validates \shortname{}'s strong ability in following instruction to complete biomedical tasks, when clear instructions are provided (\eg, yes or no).
Third, for open-set questions,  \shortname{} achieves SoTA on SLAKE, while its performance is limited on other datasets, especially compared with existing methods. This is perhaps because the open-set biomedical questions can be ambiguous without constraining their excepted answer options. 
 
\paragraph{Ablation Studies.} To study the impact of our curated instruction data and hyper-parameters in the training pipeline, we report the performance of different model variants in Table~\ref{tab:medical_vqa_model_performance} (b). Several findings are confirmed: 
$(i)$ \shortname{} consistently outperforms LLaVA by a large margin, indicating the effectiveness of our biomedical domain-specific adaptation. The performance gaps on zero-shot are larger than that in fine-tuned settings, showing that \shortname{} is clearly a better option than LLaVA when deploying one model for various scenarios in the wild.
$(ii)$ Training longer in Stage 1 improves zero-shot transfer, but Stage 1 alone is not sufficient, because the single image captioning instruction in Stage 1 may encourage the model to lose its ability in follow diverse instructions.
$(iii)$ Instruction-following data in Stage 2 is critical, and the performance is generally improved, when the instruct data amount increases from 10K to 60K. The 60K-IM data provides the best averaged zero-shot and fine-tuned performance, respectively, validating the effectiveness of considering inline mention as external knowledge in data creation.
$(iv)$ Fine-tuning longer on downstream datasets till 9 epochs benefits the performance, especially on checkpoints with 3-epoch training in Stage 2. Increasing language model size from 7B to 13B improves the overall zero-shot performance and fine-tuned performance. We suggest practitioners to choose the appropriate quality-cost trade-off, by referring to the running time in Table~\ref{tab:running_time}.

\paragraph{Case Study I: Zero-shot on Chinese Questions.} 
For the \shortname{} trained on 60K-IM data, we provide Chinese questions on SLAKE dataset.
Though \shortname{} training does not include Chinese instruction-following data, we show in Table~\ref{tab:zero_shot_on_chinese_questions} that \shortname{} is able to correctly understand the Chinese questions and respond the correct answers, probably due to the multilingual knowledge learned in LLaMA/Vicuna. Existing models will fail when zero-shot transfer cross languages.

\begin{table}
  \begin{minipage}{0.99\textwidth}
\centering  
\small
\vspace{-4mm}
\scalebox{0.93}{
\begin{tabular}{c p{2.1cm} p{7.2cm}  }
\toprule
 \multicolumn{3}{l}{\bf Biomedical image from the SLAKE Bilingual dataset. \cite{liu2021slake}}  \\
\midrule
% Caption & Chest x-ray (CXR) on Day 2 of admission post-intubation (yellow line showing the level of the endotracheal tube). Rapidly worsening ground-glass opacities are seen throughout the lungs with relative subpleural sparing (red arrows). \cite{bharat2020vaping}
% &  \includegraphics[height=3.5cm]{figures/example_imgs/visual_chat_example.jpg} \\
\multirow{3}{*}{ \includegraphics[height=2.5cm]{figures/example_imgs/xmlab280_source.jpg.jpg}  }

& 
User &  这张图片的成像方式是什么?\\
& \shortname{} & The image is an abdominal computed tomography (CT) scan, which is a cross-sectional view of the patient's abdomen. The image is taken in the portal phase, which means it is captured after the administration of contrast material to enhance the visibility of certain structures within the abdomen.
\\
& Answer & CT
\\
\midrule
\multirow{3}{*}{ \includegraphics[height=2.5cm]{figures/example_imgs/xmlab511_source.jpg.jpg}  }
& 
User & 这张图片展示的是核磁共振的哪种类型?  \\
&  LLaVA &  The image shows a typical appearance of a nodular lesion on T1-weighted MR images. The lesion appears hyperintense, which means it is brighter than the surrounding tissue. This can be indicative of various types of lesions, and further evaluation would be needed to determine the specific nature and cause of the lesion.  \\
& Answer &  T1  \\
\bottomrule
\end{tabular}
}
\vspace{1mm}
\captionof{table}{Examples on zero-shot on Chinese questions. Though \shortname{} training data itself does not include Chinese, it is able to correctly understand the Chinese questions.}
% briefly say what to notice.
% The prompt requires image understanding.}  

\vspace{-5mm}
\label{tab:zero_shot_on_chinese_questions}  
  \end{minipage}
\end{table}

\end{CJK*}

\section{Conclusions}
\label{sec:conclusions}

We present \shortname{}, a large language-and-vision model for the biomedical domain.
To create this model, we create high-quality biomedical language-image instruction-following dataset using a self-instruct approach to build a data curation pipeline using language-only GPT-4 and external knowledge.
\shortname{} demonstrates strong excellent chat abilities with domain knowledge, and outperforms previous supervised SoTA on  three \ac{VQA} datasets on certain metrics with subsequent fine-tuning.

% To create high quality medical language-image instruction-following data, we leverage self-instruct to build a data curation pipeline using language-only GPT-4 and inline mentioned external knowledge.
% We train \shortname{}, a multimodal model to assist humans on various biomedical tasks, which exhibits excellent visual chat experience in the biomedical domain.
% When zero-shot transferring to established biomedical VQA datasets, \shortname{}'s zero-shot performance is superior than previous SoTA on the closed set questions. 
% With fine-tuning, it pushes new SoTA by a significant margin. 

While we believe that \shortname{} represents a significant step towards building a useful biomedical visual assistant, we note that \shortname{} is limited by hallucinations and weak in-depth reasoning common to many \acp{LMM}. Future work is directed toward improving quality and reliability. 
% needed to improve its quality and reliability.

% \textbf{Acknowledgements.} 
% We thank Xuehai He for valuable discussions on Path-VQA dataset.
% We thank the LLaMA team for giving us access to their models, and open-source projects, including Alpaca and Vicuna.

\bibliography{egbib}
\bibliographystyle{plain}

\medskip

%%%%%%%%%%%%%%%%%%%%%%%%%%%%%%%%%%%%%%%%%%%%%%%%%%%%%%%%%%%%

\clearpage
\appendix

\section{Data}
\label{sec:data}

\paragraph{Instructions for brief image description.}
The list of instructions used to briefly describe the image content are shown in Table~\ref{tab:concise_describe_instructions}. They present the same meaning with natural language variance.

\begin{table*}[h!]\centering

\begin{minipage}{0.99\columnwidth}\vspace{0mm}    \centering
\begin{tcolorbox} 
    \centering
    \small
     \hspace{-6mm}
\begin{itemize}[leftmargin=7.5mm]
\setlength{\itemsep}{2pt}
\item "Describe the image concisely."
\item "Provide a brief description of the given image."
\item "Offer a succinct explanation of the picture presented."
\item "Summarize the visual content of the image."
\item "Give a short and clear explanation of the subsequent image."
\item "Share a concise interpretation of the image provided."
\item "Present a compact description of the photo's key features."
\item "Relay a brief, clear account of the picture shown."
\item "Render a clear and concise summary of the photo."
\item "Write a terse but informative summary of the picture."
\item "Create a compact narrative representing the image presented."
\end{itemize}

\end{tcolorbox}
    
\vspace{-2mm}
\caption{The list of instructions for brief image description.}
    \label{tab:concise_describe_instructions}
\end{minipage}
\end{table*}

\paragraph{Instructions for detailed image description.}
The list of instructions used to describe the image content in detail are shown in Table~\ref{tab:detailed_describe_instructions}. They present the same meaning with natural language variance.

\begin{table*}[h!]\centering

\begin{minipage}{0.99\columnwidth}\vspace{0mm}    \centering
\begin{tcolorbox} 
    \centering
    \small
     \hspace{-6mm}
\begin{itemize}[leftmargin=7.5mm]
\setlength{\itemsep}{2pt}
    \item "Describe the following image in detail"
    \item "Provide a detailed description of the given image"
    \item "Give an elaborate explanation of the image you see"
    \item "Share a comprehensive rundown of the presented image"
    \item "Offer a thorough analysis of the image"
    \item "Explain the various aspects of the image before you"
    \item "Clarify the contents of the displayed image with great detail"
    \item "Characterize the image using a well-detailed description"
    \item "Break down the elements of the image in a detailed manner"
    \item "Walk through the important details of the image"
    \item "Portray the image with a rich, descriptive narrative"
    \item "Narrate the contents of the image with precision"
    \item "Analyze the image in a comprehensive and detailed manner"
    \item "Illustrate the image through a descriptive explanation"
    \item "Examine the image closely and share its details"
    \item "Write an exhaustive depiction of the given image"
\end{itemize}

\end{tcolorbox}
    
\vspace{-2mm}
\caption{The list of instructions for detailed image description.}
    \label{tab:detailed_describe_instructions}
\end{minipage}
\end{table*}

\clearpage
\section{Prompts}
\label{sec:prompts}

\begin{figure}[!ht]
\begin{AIbox}{Prompting GPT-4 to generate medical visual instruction-following data}

\VarSty{messages} = [
            \{\var{"role":"system", "content":} """You are an AI assistant specialized in biomedical topics.\\
            
You are provided with a text description (Figure Caption) of a figure image from a biomedical research paper. In some cases, you may have additional text (Figure Context) that mentions the image. Unfortunately, you don't have access to the actual image.\\

Below are requirements for generating the questions and answers in the conversation:
\begin{itemize}[noitemsep,topsep=0pt,leftmargin=20pt]
    \item[-] Avoid quoting or referring to specific facts, terms, abbreviations, dates, numbers, or names, as these may reveal the conversation is based on the text information, rather than the image itself. Focus on the visual aspects of the image that can be inferred without the text information.
    \item[-] Do not use phrases like "mentioned", "caption", "context" in the conversation. Instead, refer to the information as being "in the image."
    \item[-] Ensure that questions are diverse and cover a range of visual aspects of the image.
    \item[-] The conversation should include at least 2-3 turns of questions and answers about the visual aspects of the image.
    \item[-] Answer responsibly, avoiding overconfidence, and do not provide medical advice or diagnostic information. Encourage the user to consult a healthcare professional for advice."""\}
\end{itemize}
]
        
\For{ \VarSty{sample} in   \VarSty{fewshot\_samples}}{
     \var{\VarSty{messages}.append(\{"role":"user", "content":\VarSty{sample[`context']}\})} \; \\
     \var{\VarSty{messages}.append(\{"role":"assistant", "content":\VarSty{sample[`response']}\} ) } \;
     }  
\var{\VarSty{messages}.append(\{"role":"user", "content":\VarSty{query}\})}
\end{AIbox}
\caption{\VarSty{messages} we use to prompt GPT-4 to generate medical visual instruction-following data. Manually curated few-shot examples are included in the prompt, where each example has input \VarSty{sample[`context']} and output \VarSty{sample[`response']}. Please see Figure \ref{fig:few-shot-ex} for one of the few-shot examples.} \label{fig:prompt}
\end{figure}

\begin{figure}[!ht]
\begin{AIbox}{One of the few-shot examples used in the prompt}
{\bf Example input:}\\
Figure Caption:\\
Figure 2: Chest X-ray. Cardiomegaly with diffuse bilateral interstitial infiltrates and a right-sided pleural effusion.\\

Figure Context:\\
\quad - His respiratory status continued to worsen and on hospital day number 14 , he required intubation and mechanical ventilation for hypoxemic respiratory failure ( Figure 2 ) .

\tcbline
{\bf Example output:}\\
\parbox{1\textwidth}{
\begin{alltt}
User: Which side of the lung has a pleural effusion?\\

Assistant: The right side of the lung has a pleural effusion.\\

User: According to the CT image, does the patient has cardiomegaly?\\

Assistant: The image appeares to be a chest x-ray. However, it indicates that the patient has cardiomegaly, which means an enlarged heart. Additionally, there are diffuse bilateral interstitial infiltrates and a right-sided pleural effusion present in the image.\\

User: Additional Info: The patient's respiratory status continued to worsen and on hospital day number 14, requiring intubation and mechanical ventilation for hypoxemic respiratory failure.\\

Assistant: Thank you for providing additional information about the patient's condition. Based on the information you provided, it appears that the patient's respiratory status continued to worsen and on hospital day number 14, they required intubation and mechanical ventilation for hypoxemic respiratory failure. This suggests that the patient's lung function has significantly declined, and they may be experiencing respiratory distress. It is important for the medical team to closely monitor the patient's condition and provide appropriate treatment to manage their symptoms and improve their lung function.
\end{alltt}}
\end{AIbox}
\caption{One of the few-shot examples used in our prompt to construct medical visual instruction-following data. ``Figure Context`` in the input provides additional context by including sentences from the corresponding PubMed paper that mention the figure.} \label{fig:few-shot-ex}
\end{figure}

\begin{table}
  \begin{minipage}{0.99\textwidth}
\centering  
\vspace{-4mm}
\scalebox{0.88}{
\begin{tabular}{l p{12.5cm} }
\toprule
 \multicolumn{2}{l}{\bf Visual input example, Biomedical image:}  \\
\midrule
&  \includegraphics[height=3.5cm]{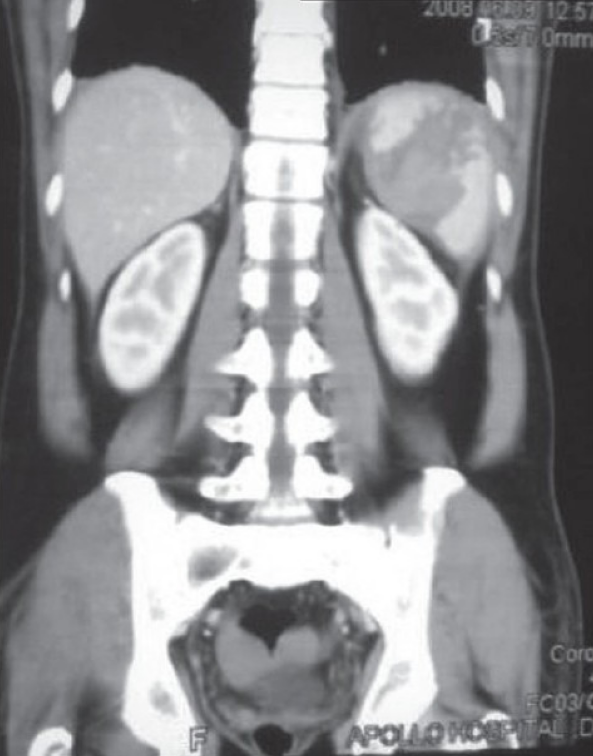} \\
% & {\tiny Source:~\url{https://www.barnorama.com/wp-content/uploads/2016/12/03-Confusing-Pictures.jpg}} \\
User & What is this image about?\\
\shortname{} & The image is a coronal view of a computed tomography (CT) scan of the abdomen and pelvis. It provides a detailed view of the internal organs and structures within the abdominal and pelvic cavities.
\\
User & What is abnormal about this image? \\
\shortname{} & The abnormality in this image is the presence of a large left-sided pelvic mass. This mass could be due to various reasons, such as a tumor, cyst, or other growth. Further evaluation and tests would be needed to determine the exact cause and appropriate treatment for this abnormality.
\\
\midrule
User & What is this image about? \\
LLaVA &  The image is a medical illustration showing a person's internal anatomy, especially focusing on their large and small intestines, stomach, liver and kidneys. \textcolor{red}{The organs are highlighted in different colors}, making it easy to identify each organ \\
User & What is abnormal about this image? \\
LLaVA &  \textcolor{red}{As a visual AI assistant, I cannot determine the specific abnormality on the chest X-ray based on the provided image description. The image description mainly focuses on the organs and their colors...} \\
\bottomrule
\end{tabular}
}
\vspace{1mm}
\captionof{table}{Example comparison of medical visual chat and reasoning capabilities. }
% briefly say what to notice.
% The prompt requires image understanding.}  
\vspace{-5mm}
\label{tab:visual_chat_example1_appendix}  
  \end{minipage}
\end{table}

\end{document}